\documentclass[a4paper,conference]{IEEEtran}

\usepackage{times}
\usepackage{soul}
\usepackage{url}
\usepackage[hidelinks]{hyperref}
\usepackage[utf8]{inputenc}
\usepackage[small]{caption}
\usepackage{graphicx}
\usepackage{amsmath, bm}
\usepackage{amsthm}
\usepackage{booktabs}
\usepackage{algorithm}
\usepackage{algorithmic}
\usepackage{multirow}
\usepackage{amsfonts,amssymb}
\usepackage[caption=false,font=normalsize,labelfont=rm,textfont=rm]{subfig}

\ifCLASSINFOpdf
\else
\fi

\begin{document}
%
\title{Differentiable Channel Sparsity Search via \\Weight Sharing within Filters}

\author{\IEEEauthorblockN{Yu Zhao}
\IEEEauthorblockA{Xidian University\\
Xi'an, China\\
Email: zhaoyu775885@163.com}
\and
\IEEEauthorblockN{Chung-Kuei Lee}
\IEEEauthorblockA{ HiSilicon Technologies\\
Shanghai, China\\
Email: lee.chung.kuei@hisilicon.com}
}


%


\maketitle

\begin{abstract}
In this paper, we propose the differentiable channel sparsity search (DCSS) for convolutional neural networks.
Unlike traditional channel pruning algorithms which require users to manually set prune ratios for each convolutional layer, DCSS automatically searches the optimal combination of sparsities.
Inspired by the differentiable architecture search (DARTS), we draw lessons from the continuous relaxation and leverage the gradient information to balance the computational cost and metrics. 
Since directly applying the scheme of DARTS causes shape mismatching and excessive memory consumption, 
we introduce a novel technique called weight sharing within filters.
This technique elegantly eliminates the problem of shape mismatching with negligible additional resources. 
We conduct comprehensive experiments on not only image classification but also find-grained tasks including semantic segmentation and image super resolution to verify the effectiveness of DCSS.
Compared with previous network pruning approaches, DCSS achieves state-of-the-art results for image classification. 
Experimental results of semantic segmentation and image super resolution indicate that task-specific search achieves better performance than transferring slim models, demonstrating the wide applicability and high efficiency of DCSS.
\end{abstract}


%
\IEEEpeerreviewmaketitle

\section{Introduction}
Convolutional neural networks (CNNs) have achieved the state-of-the-art performance on many computer vision tasks.
However, the requirement of computational resources remains a major obstacle which prevents CNNs' deployment on mobile devices.
Various model compression techniques have been developed to balance the resource consumption and metrics of specific tasks.
As one of the most prevailing methods for model compression, 
channel pruning has the advantage of directly inheriting the runtime environment of the unpruned models, thanks to the retention of the overall structures without altering operator types or introducing extra ones.

Traditional channel pruning approach attempted to compress each layer of the network by given prune ratios \cite{he2017channel}, which relies heavily on domain expertise. 
Since exhaustive traversal of the state space is impractical, heuristic criteria for channel pruning were widely studied.
As the representatives, methods in \cite{he2019filter} and \cite{liu2017learning} distinguish the importance of different filters with specific metrics.
However, as pointed out in \cite{liu2018rethinking}, 
it is the structure of the pruned network that affects the performance instead of the weights inherited from the full-size model.
Thus, the challenge for channel pruning lies in searching the optimal combination of prune ratios. 
From this perspective, techniques for NAS could be borrowed for sparsity search and model compression.
In AMC \cite{he2018amc}, reinforcement learning was utilized to find optimal combination of sparsities for each convolutional layer.
Their experiment proved that for a given convolutional network, the redundancy of each layer varies widely and dedicated combination of sparsities can significantly improve the performance of the pruned network.

\begin{figure}[t!]
	\setlength{\abovecaptionskip}{0.cm}
	\setlength{\belowcaptionskip}{0.cm}
	\centering
	\includegraphics[width=3.4in]{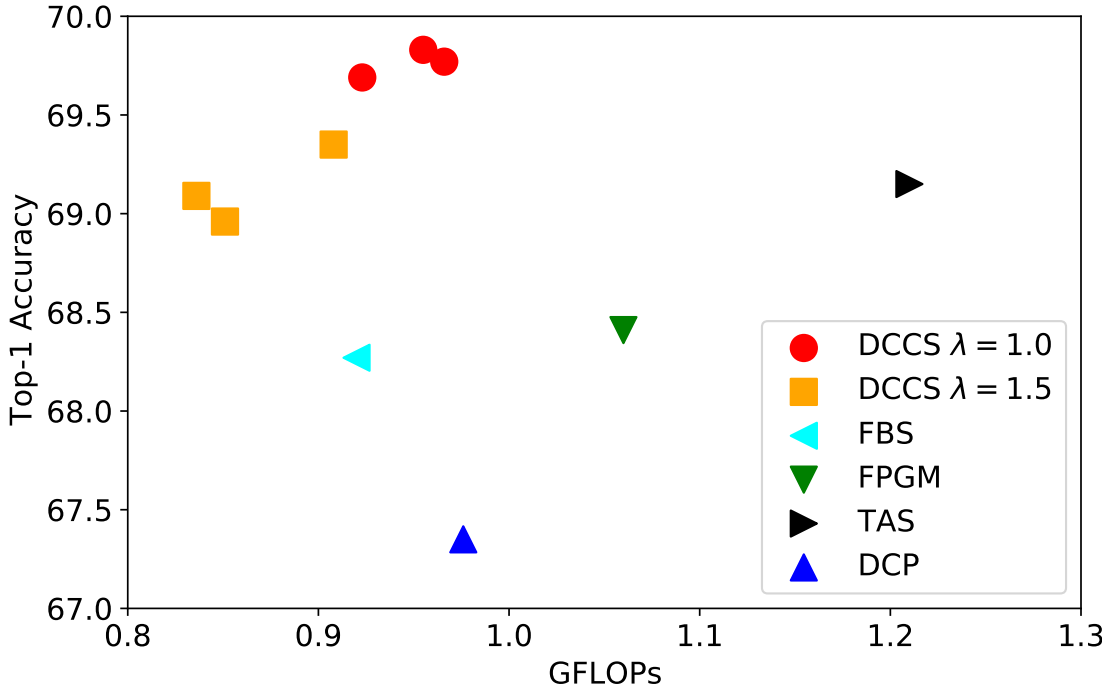}
	\vspace{0.1in}
	\caption{Top-1 accuracy of the pruned ResNet-18 on ImageNet v.s. MFLOPs.
		DCSS is performed consecutively three times for 2 different values of $\lambda$ to verify the stability.
		DCSS constantly outperforms the previous state-of-the-art methods 
			FBS\protect\cite{gao2019dynamic}, FPGM\protect\cite{he2019filter}, TAS\protect\cite{dong2019network} and DCP\protect\cite{zhuang2018discrimination}.}
	\label{fig:resnet18}
\end{figure}
\setlength{\belowcaptionskip}{0in}

While the effectiveness of applying NAS techniques on model compression has been proved, the resource consumption during searching is still prohibitive.
NAS approaches based on reinforcement learning or evolutionary algorithms require a huge amount of computational resources for exploration.
For instance, the evolutionary algorithm based method \cite{shu2019co} spent 60 GPU days searching the optimal number of filters on Cityscape dataset.
As a result, differentiable approaches were introduced to tackle the resource consumption problem of NAS.
In DARTS \cite{liu2019darts}, the authors proposed to optimize the probability distribution of each candidate operator via gradient decent.
FBNet \cite{wu2019fbnet} followed a similar procedure to jointly optimize the latency and accuracy of the network.
Experiments showed that approaches based on differentiable search are much more efficient than those based on sequential optimizations involving multi-step training and evaluation.
Despite the success in NAS, differentiable search scheme has its own difficulty in transferring to pruning tasks.
One immediate problem is the size of the search space. 
DARTS searches the possible configuration of a block within limited number of candidate operators, while the candidates for channel pruning are the possible number of channels, which may reach up to several hundreds or thousands and result in excessive memory overhead.
Besides, convolutions with distinct number of channels lead to problem of shape mismatching so that the element-wise additions of feature maps are not permitted.


In this paper, we present the differentiable channel sparsity search (DCSS) for deep convolutional neural network.
Inspired by the recent advances in differentiable architecture search, we propose to leverage weight sharing within filters to resolve the shape mismatching and excessive memory overhead problems simultaneously.
Weight sharing within filters technique allows the additions of feature maps to be reorganized as weighting each channel separately, which elegantly eliminate the problem of shape mismatching with negligible additional resources.
We conduct experiments for three types of tasks, including image classification on CIFAR-100 and ImageNet, semantic segmentation on PASCAL VOC 2012 and image super resolution on DIV2K.
For image classification, DCSS outperforms the recent state-of-the-art channel pruning methods in terms of both accuracy and FLOPs. 
Moreover, experiments for ResNet-18 on ImageNet are conducted several times to verify the stability of DCSS, and it constantly outperforms the state-of-the-art methods as illustrated in Fig.\;\ref{fig:resnet18}.
To our knowledge, we are the first to perform channel pruning technique on segmentation and super resolution tasks.
Experiments on fine-grained tasks including semantic segmentation and super resolution indicate that task-specific channel pruning achieves better performance against transferring slim models, 
which further dimonstrate the wide applicability and high efficiency of DCSS.


\section{Related Work}
\subsection{Structured Network Pruning}
Network pruning is an effective technique for model compression and inference-time reduction of deep neural networks by removing redundant weights and structures in the model.
Abundant approaches using different importance metrics and pruning unit have been proposed.
Pruning algorithms can be categorized into two categories, unstructured pruning and structured pruning.

Despite the high sparsity achieved by unstructured pruning algorithms, the inference time on real-world hardware is usually much longer than the theoretical inference time \cite{wen2016learning}. 
To make the pruned model more hardware-friendly, people attempted to delete the whole sub-structure in the convolution kernel instead of removing parameters one-by-one. 
The group Lasso was proposed to remove filters in deep neural networks \cite{wen2016learning}.
The method proposed in \cite{zhuang2018discrimination} determined the importance of each filter by investigating their impact to the loss function and then cut the least important ones out.
There are a lot more structured pruning algorithms, we refer to \cite{blalock2020state} for a more comprehensive survey.

\subsection{Sparsity Combination Search}
Experiments in AMC \cite{he2018amc} showed that the reinforcement learning based channel pruning beat other handcrafted policy by a large margin.
Following this line of development, lots of new approaches surfaced.
The genetic algorithm was leveraged to find the optimal sparsity of structure
combination for deep neural networks \cite{shu2019co}. 
Besides reinforcement learning and genetic algorithms, gradient based optimization techniques also yielded fruitful results.
Network slimming \cite{liu2017learning} used the gradient to induce sparse scaling factors for
batch-normalization layers and remove redundant filters accordingly. 
However the scaling factors it relied on do not have uniform meaning for different layers making the algorithm more heuristic.
Network slimming was further extended in \cite{huang2018data}
to more general models by adding controlling gates to the convolutional kernels. 
Besides, it replaced $L_1$-regularization by the element-wise mean of gates and presented an algorithm to determine the threshold for gate elimination. 
To incorporate differentiable search with channel pruning, the transformable architecture search (TAS) proposed the channel wise interpolation (CWI) to alleviate the problem of shape mismatching, and the number of sampling candidates was restricted to be $2$ to save memory during searching \cite{dong2019network}.
The performance of the classifier network in TAS was further boosted by knowledge distillation, which is also adopted in this work.
In \cite{dong2019network}, the gates were replaced by Gumbel-Softmax which models a probability distribution. By optimizing the distribution through gradient, a more compact model is obtained.
Differentiable markov channel pruning (DMCP) followed a similar line of thought, except that it modeled the channel pruning as as a Markov process to realize the differentiable search \cite{dcmp2020}.

\begin{figure*}[htbp]
	\setlength{\abovecaptionskip}{0.cm}
	\setlength{\belowcaptionskip}{0.cm}
	\centering
	\subfloat[]{\includegraphics[width=2.1in]{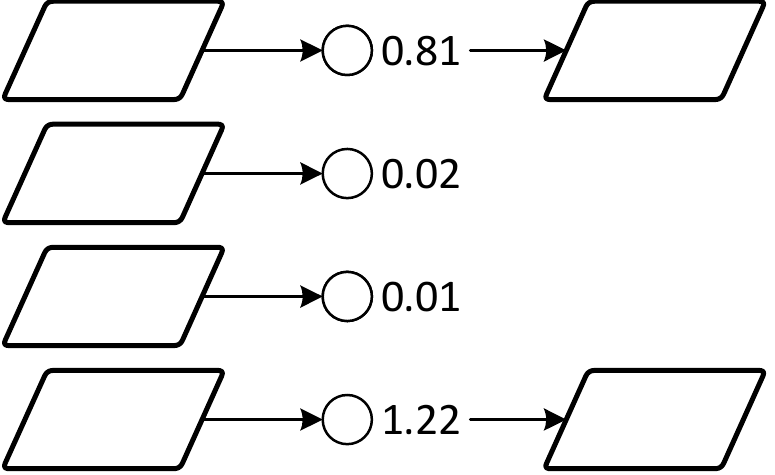}}
	\hspace{0.1in}
	\subfloat[]{\includegraphics[width=2.1in]{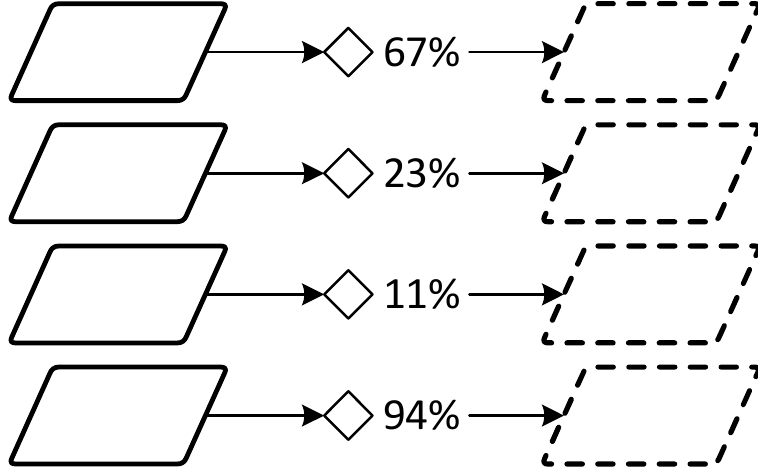}}
	\hspace{0.2in}
	\subfloat[]{\includegraphics[width=2.4in]{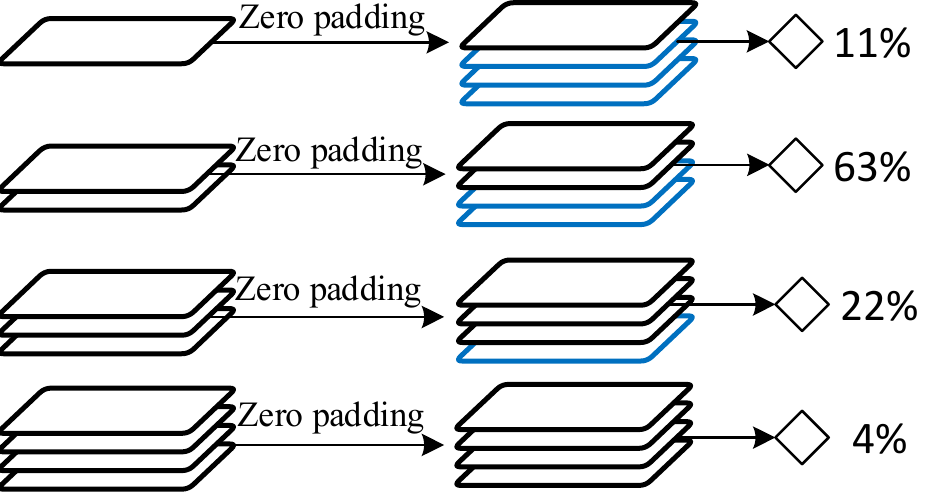}} \\
	\subfloat[]{\includegraphics[width=3.2in]{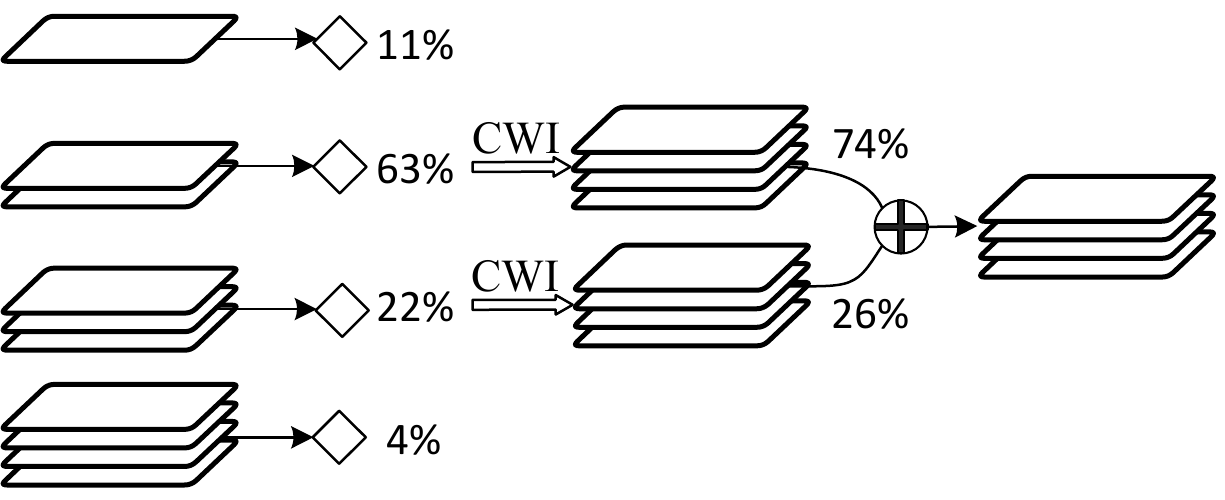}}
	\hspace{0.0in}
	\subfloat[]{\includegraphics[width=3.2in]{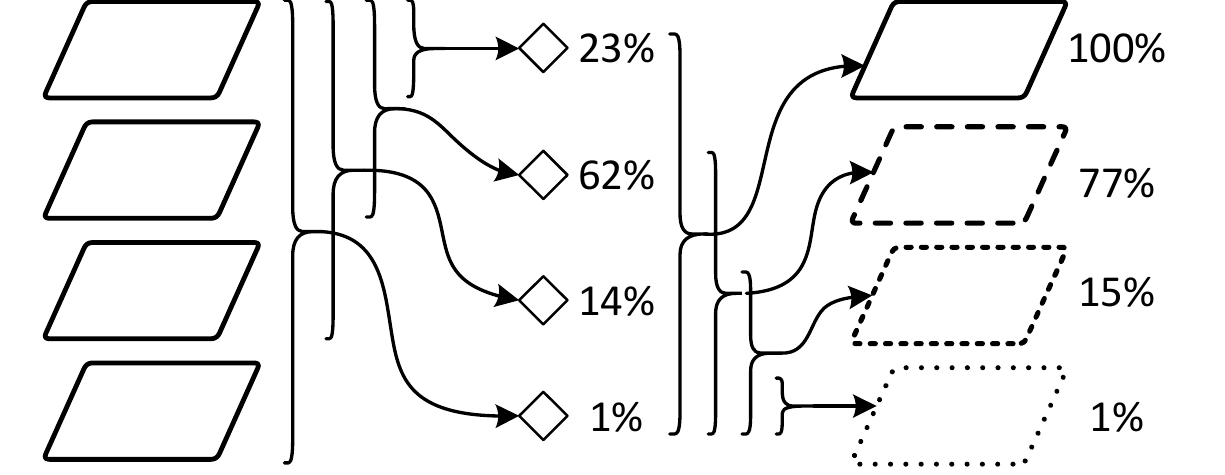}}\\
	\caption{Intuitive comparison of different pruning algorithms. (a) Network slimming. (b) Probabilistic Gate. (c) Differentiable neural architecture search. (d) Transformable architecture search. (e) Differentiable channel sparsity search.}
	\label{fig:1}
\end{figure*}

\section{Methodology}
In this section, we first formulate the problem of searching the optimal combination of sparsities and then present the basic idea of the proposed method. Then we elaborate on the optimization of the combination of sparsities under the differentiable search scheme via weight sharing within filters.

\subsection{Problem Formulation}
Our goal is to find a balance between the metrics of the model and the computational cost by searching the optimal combination of the number of channels for each convolutional layer.
We denote $\mathcal{L}_{task}$ to be the loss function of the specific task for the network to be pruned and $\mathcal{C}_{task}$ to represent the computational cost.
The problem can be formulated as
\begin{equation}
\min_{a\in \mathcal{A}}\min_{W_{a}}[\mathcal{L}_{task}(a, W_a)+\lambda\mathcal{C}_{task}(a)],
\label{eq:1}
\end{equation}
where $a$ denotes a combination of the number of channels in the search space $\mathcal{A}$, $W_a$ is the weight of the pruned model and $\lambda$ is the coefficient balancing the computational cost and the metrics of specific tasks. 

In this work, computational cost $\mathcal{C}_{task}$ is measured by $\log{(\text{FLOPs})}$.
For the $i$-th convolutional layer, the height, width and number of channel for the output feature maps are assumed to be $H_i$, $W_i$ and $C_i$, and the height, width and number of channel for the corresponding kernel are assumed to be $h_i$, $w_i$ and $n_i$.
According to \cite{molchanov2017}, the FLOPs of the $i$-th convolutional layer without bias is 
\begin{equation}
\text{FLOPs}_{i} = h_{i}w_{i}n_{i}H_{i}W_{i}C_{i}.
\label{eq:2}
\end{equation}
Let $\eta_{i} = h_{i}w_{i}n_{i}H_{i}W_{i}$, then we have $\text{FLOPs}_{i} = \eta_{i}C_{i}$.
Suppose $a\in\mathcal{A}$ is sampled from the probability distribution $\mathcal{D}$,
the optimization problem in (\ref{eq:1}) can be reformulated as searching the optimal distribution $\mathcal{D}$:
\begin{equation}
\min_{\mathcal{D}}\min_{W}\left[\mathbb{E}_{a\sim\mathcal{D}}\left(\mathcal{L}_{task}(a,
W_a)\right)+\lambda\log\left(\sum_{i}\eta_i\mathbb{E}_{\mathcal{D}}[C_i]\right)\right].
\label{eq:3}
\end{equation}

\subsection{Differentiable Channel Sparsity Search}
A rather straight forward way to solve (\ref{eq:3}) is to extend the method in \cite{liu2017learning} by replacing the scaling factors with probabilities, as shown in Fig.\:\ref{fig:1}\,(b).
However, it is basically identical with network slimming and suffers the similar weakness, such as easily trapped in poor local minima.

Another approach is to extend the idea of \cite{wu2019fbnet} and consider convolutions with distinct number of channels as different operators.
Suppose that the set of $K$ candidate number of filters is $\{c_{k}\}$, $k=1,\ldots,K$ for the $i$-th convolutional layer.
For the sake of clarity, the layer's index is omitted.
We denote $o_{k}(\cdot)$ as the convolution operator with $c_{k}$ filters.
Each filter keeps all the other remaining settings identical with the $i$-th layer of the unpruned original network, of which the input and output are denoted as $x$ and $y$, respectively.
Then the discrete selection of particular operation is relaxed by the expectation over all candidates, and the output of the expectation is expressed by
\begin{equation}
\hat{y} = \sum_{k} \mathbb{P}(c=c_{k})o_{k}(x).
\label{eq:4}
\end{equation}

A natural choice to model the probability $\mathbb{P}(c=c_{k})$ is the softmax function.
However, as \cite{xie2018snas} pointed out, modeling probability distribution with softmax would suffer from the high variance of likelihood ratio trick.
The Gumbel Softmax parameterized by $\bm{\theta}$ is a better choice
\begin{equation}
\mathbb{P}(c=c_{k}) = \frac{exp((\theta_{k}+g_{k})/\tau)}{\sum_{j}{exp((\theta_{j}+g_{j})/\tau)}}
\label{eq:5}
\end{equation}
where $\bm{\theta{}}\in\mathbb{R}^{K}$ is the vector of controlling gates, $g_{k}\sim{}Gumbel(0, 1)$ is noise obeying the Gumbel distribution, and $\tau$ is the softmax temperature.
Under such settings, the discrete selection is relaxed to probabilistic sampling, which transforms the search problem into the optimization of gates.
However, the addition operation is not allowed for feature tensors with different shapes in channel dimension, thereby prohibiting the direct application of differentiable search methods based on (\ref{eq:4}).

A simple solution for this problem is zero-padding along channel dimension, as illustrated in Fig.\:\ref{fig:1}\,(c).
Suppose the original convolution has $64$ output channels and only
$3$ candidate number of channels, say $32$, $48$ and 64, are considered.
To implement additions, both the $32$-channel and $48$-channel tensors would be zero-padded up to $64$ channels.
TAS proposed the CWI to dynamically match the channel dimension of feature maps.
As illustrated in Fig.\;\ref{fig:1}\,(d), the outputs of the two most likely operators are expanded to the same number of channels by CWI and then summed up together.
The drawback of these two approaches is apparent.
As shown in Fig.\;\ref{fig:1}\,(c) and Fig.\;\ref{fig:1}\,(d), each path of the selections requires an independent convolutional kernel and yields an output tensor with full number of channels.

General cases get worse.
Suppose the $i$-th convolutional layer has $n$ filters in the
original model.
There will be $n$ independent paths corresponding to $c_{k}=k$ for $k=1,\cdots,n$,
which cost roughly 
$n$ times more computational resources than the original model.
TAS circumvented this by restricting the number of the sampling candidates to be $2$.
Nevertheless, TAS still has to cost nearly twice the memory of the unpruned model.
Besides, the CWI which is implemented by adaptive average pooling may sacrifice a little precision.

\subsection{Weight Sharing within Filters}
In order to overcome the aforementioned resource exhaustion obstacle, we developed a novel weight sharing technique to the interior of convolutional kernel. 
We denote the convolution operation by $*$ and the original convolution by $y=W*x$.
Since the probability is controlled by the gates, for ease of searching, we preset the candidates with large enough distinctions.
For simplicity, we split the filters of the unpruned convolutional layer into groups, and collect the candidates by incrementally merging the groups.
Moreover, we use $W_k$ to represent a kernel whose weights of the first $c_{k}$ filters are the same with $W$ while the remaining weights are zeroes.
In other words, if we let the original kernel be $W$ and the $k$-th filter
of $W$ be $W[k]$, then
\begin{align}
\begin{aligned}
W_{k}[c] = \begin{cases}
W[c] &, c \leq c_{k};  \\
\quad 0 &, \hbox{otherwise}.
\end{cases}
\label{eq:6}
\end{aligned}
\end{align}
Similarly, we define $y[c]$ to be the value of the $c$-th channel of $y$,
then the convolutions $y_{k}= o_k(x)=W_k*x$ yields
\begin{align}
\begin{aligned}
&y_{k}[c] = \begin{cases}
y[c] &, c\leq c_{k};  \\
\quad 0 &, \hbox{otherwise}.
\end{cases}
\label{eq:7}
\end{aligned}
\end{align}
Finally, each slice of the output of (\ref{eq:4}) can be expressed by
\begin{align}
\begin{aligned}
\hat{y}[c_{k-1}:c_{k}]
&= \sum_{j=k}^{K} \mathbb{P}(c=c_{j})y_{j}[c_{k-1}:c_{k}]  \\
&= \left(\sum_{j=k}^{K} \mathbb{P}(c=c_{j})\right)y[c_{k-1}:c_{k}]
\label{eq:8}
\end{aligned}
\end{align}
The addition of the features weighted by softmax is transformed into feature weighting along the channel dimension by (\ref{eq:8}).
To make it more clear, we exhibit the whole process in Fig.\;\ref{fig:1}\,(e).
The left part of the figure demonstrates what $y_{k}=W_k*x=o_k(x)$ actually represent.
The diamonds in the figures symbols $\mathbb{P}(c=c_j)$ and the right part displays the feature weighting derived in (\ref{eq:8}).

There are two major advantages of such settings.
The first and most obvious one is about the efficiency.
Since (\ref{eq:4}) is realized as weighting each channel of the feature, the additional computation cost and memory requirement are negligible and no intermediate variables are required.
The second advantage is to avoid redundant information.
Suppose $c_1, c_2, c_3$ are the candidate numbers of channels with $c_1<c_2<c_3$, 
the information contained in $o_{1}(x)$ and $o_{2}(x)$ could be mutually exclusive without weight sharing. 
It leads to a possibility that all the information contained in $o_{3}(x)$ is covered by the union of $o_{1}(x)$ and  $o_{2}(x)$.
Under such circumstance, the algorithm will simply minimize $\mathbb{P}(c=c_3)$.
With weight sharing, the information in the feature of $o_{3}(x)$ is strictly more than $o_{2}(x)$ and $o_{2}(x)$ is strictly more than $o_{1}(x)$.
It ensures that the value of $\mathbb{P}(c=c_3)$ is decided by the usefulness of the additional information from $o_{3}(x)$, which suits our purpose much better.

\subsection{Optimization and Model Extraction}
We use a similar optimization scheme as \cite{wu2019fbnet} to search the optimal structure.
The full training set is split into weight training set denoted as $\mathbb{D}_{train}$ and validation set denoted as $\mathbb{D}_{val}$.
The validation set $\mathbb{D}_{val}$ is used for the optimization of the controlling gates.
The overall optimization has $2$ stages.
The first stage is warming-up with zero initialized gates, meaning the initial probability for each selection is equal.
The second stage is searching.
We found that the fully optimized warming-up model provides good initials for the convergence during searching.
For searching stage, weights are updated with fixed gates using data from $\mathbb{D}_{train}$, and gates are optimized with fixed weights using data from $\mathbb{D}_{val}$.
The weights and gates are optimized alternately to minimize the loss function in (\ref{eq:3}) in each step.

Because the probability for each candidate does not actually represent the contribution, 
it is unwise to follow the common way for categorical selection that takes the item with the largest probability.
Indeed, as \cite{liu2018rethinking} pointed out, it is the size of
the total chosen groups that really matters, meaning what we are searching is the number of channels but not specific filter.
As a result, our actual targets are $\mathbb{E}_{\mathcal{D}}[c_i]$ from (\ref{eq:3}) for
all layers, which can be approximated by
\begin{equation}
\mathbb{E}_{\mathcal{D}}[c] \sim \sum_{k=1}^K c_{k}\cdot\mathbb{P}(c=c_{k}).
\label{eq:9}
\end{equation}
After evaluating $\mathbb{E}_{\mathcal{D}}[c]$, the pruned model is extracted with the acquired number of channels for each convolutional layer.


\section{Experiments}
In this section we present experimental results to demonstrate the effectiveness and feasibility of the techniques described above.
We evaluate DCSS on image classification, semantic segmentation and super resolution tasks.

\subsection{Benchmark Datasets and Common Settings}
CIFAR-100 contains 60K $32\times{}32$ RGB images (50K for trainging and 10K for testing) belonging to 100 categories.
We randomly split the training set with 45K for weight training, and the remaining 5K for gate searching.

ImageNet \cite{imagenet} is composed of 1.28 million training samples and 50K testing images from 1K classes.
We take 10K images in the training set for gate searching and the remaining for weight training.

PASCAL VOC 2012 \cite{pascal} dataset, which constains 20 foreground object classes and one background class, is used for semantic segmentation.
We augment the dataset following the description in \cite{chen2017rethinking}, which result in $10,582$ training samples.
We randomly pick 1K images for gate searching and use the left images for weight training.
And the validation set containing $1,449$ images is used to evaluate the performance.

We use DIV2K \cite{div2k}, a dataset comprised of 1000 2K resolution images (800 for training), for image super resolution. We generate roughly 12.8K 48$\times$48 patches from the training set.
About 10\% of the patches are picked to optimize the gates and the rest to optimize the weights.

We use stochastic gradient decent (SGD) with a momentum of 0.9 to optimize weights and Adam optimizer is used to optimize gates.
The learning rate for optimization during warming-up and searching follows the original training process of the full-size model, except that the initial learning rate for searching is $10$ times reduced. 
Adam optimizer adopts $\frac{1}{10}$ of SGD's learning rate during searching.
The number of epochs for both warming-up and searching is shortened to less than $50\%$ of the normal training process.
The channels of each layer are split into $8$ groups with nearly identical sizes.
The temperature $\tau$ anneals exponentially from $10$ to $0.1$.
During fine-tuning stage, the slim model takes identical learning strategy as training the unpruned model. 
The coefficient $\lambda$ is chosen from $\{0.5, 1.0, 1.5\}$ without fine-tuning for all tasks, except that for super resolution $\lambda$ is chosen to be $0.01$.

\subsection{Baseline Algorithms}
We compare our method with the following pruning algorithms to evaluate the performance of DCSS for image classification.
\textbf{AMC} \cite{he2018amc} is a reinforcement learning based sparsity search algorithm.
Other prune ratio deciding strategies like greedy algorithm (\textbf{DCP} \cite{zhuang2018discrimination}), soft selection (\textbf{SFP} \cite{he2018soft}) and differentiable search algorithm (\textbf{TAS} \cite{dong2019network}, \textbf{DMCP} \cite{dcmp2020}) are also included for comparison.
For the sake of completion, some other state-of-the-art channel pruning algorithms including \textbf{FPGM} \cite{he2019filter}, \textbf{FBS} \cite{gao2019dynamic},  and \textbf{PFS} \cite{wang2019pruning} are also listed.
\textbf{SlimConv} \cite{SlimConv2021}, which was designed to reform feature channels to improve the quality of representations, is also compared with out proposed method.

\subsection{Image Classification on CIFAR-100}
\textbf{Model description and settings}.
We train ResNet-20, 32 and 56 with 16 base-channels on image classification for CIFAR-100.
For normal training, the training strategy is kept the same as TAS \cite{dong2019network}.
Both the warming-up and searching takes $150$ epochs.
The learning rate starts from $0.1$ and is divided by $10$ at the $100$-th epoch for warming-up.
During searching, the learning rate starts from $0.01$ and is divided by $10$ at the $50$-th and $100$-th epoch.
We choose $\lambda$ to be $1.0$ for this experiment.

\begin{table}[htbp]
	\centering
	\begin{tabular}{c c c c c c}
		\toprule[1pt]
		&Method & Acc.\;(\%) & Drop\;(\%) & FLOPs & Pr.\;(\%) \\
		\toprule
		\parbox[t]{2mm}{\multirow{4}{*}{\rotatebox[origin=c]{90}{ResNet-20}}}& FPGM & $66.86$ & $0.76$ & $24.3$M &  $42.2 $\\                                                                            
		& SFP & $64.37$ & $ 3.25$ & $24.3$M &  $42.2$\\
		& TAS & $68.90$ & $-0.21$ & $22.4$M &  $45.0$\\
		\cmidrule{2-6}
		&DCSS & $69.17$ & $ 0.00$ & $17.9$M &  $56.4$\\
		\midrule
		\parbox[t]{2mm}{\multirow{4}{*}{\rotatebox[origin=c]{90}{ResNet-32}}}& FPGM & $68.52$ & $1.25$ & $40.3$M &  $41.5 $\\
		& SFP & $68.37$ & $ 1.40$ & $40.3$M &  $41.5$\\
		& TAS & $72.41$ & $-1.80$ & $42.5$M &  $38.5$\\
		\cmidrule{2-6}
		&DCSS & $70.22$ & $ 0.07$ & $34.3$M &  $50.6$\\
		\midrule
		\parbox[t]{2mm}{\multirow{4}{*}{\rotatebox[origin=c]{90}{ResNet-56}}}& FPGM & $69.66$ & $1.75$ & $59.4$M &  $52.6 $\\
		& SFP & $68.79$ & $2.61$ & $59.4$M &  $52.6$\\
		& TAS & $72.25$ & $0.93$ & $61.2$M &  $51.3$\\
		\cmidrule{2-6}
		&DCSS & $72.33$ & $1.14$ & $53.0$M &  $57.9$\\
		\bottomrule[1pt]
	\end{tabular}
	\caption{Comparison of different channel pruning algorithms for ResNet-20, 32, and 56 on CIFAR-100. "Pr." denotes prune ratio.}
	\label{tab:1}
\end{table}

\textbf{Performance study and comparison}. 
The results of the experiment are summarized in Table \ref{tab:1}.
The "Pr." denotes prune ratio in all the following tables.
We compare the performance of the pruned model with those from several state-of-the-art pruning methods.
DCSS achieves more than $50\%$ prune ratios for all the three networks with very little accuracy drop.
Compared with the other methods, DCSS obtains the highest accuracy and the lowest FLOPs except for the ResNet-32 with TAS, which showed $1.8\%$ accuracy gain with $38.5\%$ FLOPs pruned.


\subsection{Image Classification on ImageNet}
\textbf{Training and searching settings}.
In this section, we show the performance of DCSS for pruning ResNet-18 and ResNet-50 on ImageNet.
We still take the same training strategy as TAS. 
The warming-up and searching process both take $50$ epochs.
The learning rate starts from $0.1$ and is divided by $10$ at the $30$-th epoch for warming-up.
During searching, the learning rate starts from $0.01$ and is divided by $10$ at the $30$-th epoch.
We perform DCSS for ResNet-18 consecutively several times with $\lambda$ being $1.0$ and $1.5$ to verify the stability of our approach.
We choose $\lambda$ to be $0.5$ for ResNet-50.


\begin{table}[htbp]
	\centering
	\begin{tabular}{c c c c c c}
		\toprule[1pt]
		&Method & Acc.\;(\%) & Drop\;(\%) & GFLOPs & Pr.\;(\%) \\
		\toprule
		\parbox[t]{2mm}{\multirow{5}{*}{\rotatebox[origin=c]{90}{ResNet-18}}}& FPGM & $68.41$ & $1.87$ & $1.06$ & $41.8$\\
		& DCP & $67.35$ & $2.39$ &  $0.98$ & $46.1$ \\
		& FBS & $68.27$ & $2.54$ &  $0.92$ & $49.5$ \\
		& TAS & $69.15$ & $1.50$ &  $1.21$ & $33.3$ \\
		\cmidrule{2-6}
		& DCSS& $69.83$ & $1.10$ &  $0.96$ & $47.4$ \\
		\midrule
		\parbox[t]{2mm}{\multirow{8}{*}{\rotatebox[origin=c]{90}{ResNet-50}}}& FPGM & $75.50$ & $0.65$ & $2.36$ & $42.2$\\
		& DCP& $74.95$ & $1.06$ & $1.82$ & $55.6$\\
		& PFS& $75.6$ & $1.6$ & $2.00$ & $51.1 $\\		
		& AutoSim & $75.6$ & $0.5$ & $2.00$ & $51.1$\\
		& TAS& $76.20 $ & $1.26 $ & $2.31$ & $43.5$\\
		& DCMP& $76.2 $ & $0.4 $ & $2.2$ & $43.5$\\		
		& SlimConv & $75.52 $ & $0.63$ & $1.88$ & $54.0$\\				
		\cmidrule{2-6}		
		& DCSS& $76.80 $ & $0.29$ & $1.89$ & $53.8$\\
		\bottomrule[1pt]
	\end{tabular}
	\caption{Comparison of different pruning algorithms for ResNet-18 and ResNet-50 on ImageNet.}
	\label{tab:3}
\end{table}

\textbf{Performance study and comparison}.
The results of the experiments for pruning ResNet-18 and ResNet-50 are shown in Table \ref{tab:3}.
Note that for ResNet-18 in Table \ref{tab:3}, the result is chosen from Fig.\;1 with $\lambda=1.0$.
As can be observed, DCSS outperforms the previous state-of-the-art methods in terms of the both the accuracy and prune ratio.
Besides, DCSS also achieves the smallest accuracy drops.
The results of multiple executions of DCSS shown in Fig.\;1 demonstrate that the proposed DCSS is rather stable and gets closer to the Pareto front for the channel pruning problems.


\subsection{Semantic Segmentation on PASCAL}
\textbf{Training and searching settings}.
We adopt the fully convolutional network (FCN) \cite{long2015fully} for semantic segmentation.
Both ResNet-18 and ResNet-50 are chosen as the backbones for FCN-32s and FCN-16s.
FCNs are trained $500$ epochs from scratch without pre-training on ImageNet.
For ResNet-18 based and ResNet-50 based FCN, the training images are cropped to $224\times{}224$ and $513\times{}513$, respectively.

\begin{table}[htbp]
	\centering
	\begin{tabular}{c c c c c c}
		\toprule[1pt]
		&\multirow{2}{*}{Method} & \multicolumn{2}{c}{FCN-32s} & \multicolumn{2}{c}{FCN-16s}\\
		&  & mIoU(\%) & Pr.(\%) & mIoU(\%) & Pr.(\%)\\
		\toprule
		\parbox[t]{2mm}{\multirow{4}{*}{\rotatebox[origin=c]{90}{ResNet-18}}}& Baseline & $51.91 $ & - &  $53.61 $ & - \\
		& Uniform & $50.98 $ & $45.9 $ &  $52.06$ & $45.9 $ \\
		& Transfer & $50.45$ & $46.3$ &  $51.69$ & $46.3 $ \\
		& DCSS & $51.15 $ & $47.3$ &  $52.18$ & $46.1$ \\
		\midrule
		\parbox[t]{2mm}{\multirow{4}{*}{\rotatebox[origin=c]{90}{ResNet-50}}}& Baseline & $57.86$ & - &  $58.91$ & - \\
		& Uniform & $56.37 $ & $55.6 $ &  $56.72$ & $55.6$ \\
		& Transfer & $56.85$ & $55.6$ &  $57.32$ & $55.6$ \\
		& DCSS & $57.80$ & $55.5$ &  $57.70 $ & $55.6$ \\
		\bottomrule[1pt]
	\end{tabular}
	\caption{Comparison of different methods for FCN on PASCAL VOC 2012. "Uniform" denotes uniform pruning, and "Transfer" denotes transferring pruned model from ImageNet classification.}
	\label{tab:7}
\end{table}

\textbf{Performance study and comparison}.
The results are shown in Table \ref{tab:7}.
We adopt both the uniform pruning method and transferring pruned models from the ImageNet classification task from previous section as comparisons.
According to Table \ref{tab:7}, models from DCSS have better mean intersection-over-union (mIoU) than those from other methods with nearly the same prune ratio.
Uniform pruning is rarely the best strategy, which meets our expectation.
We need to emphasize that channel pruning should be performed for each specific task, since each pruning pattern is shrunken closely with the problem with little redundancy, which explains the reason why transferred models does not have the best performance.
As aforementioned, computer vision tasks, including detection and segmentation, consume large memory for training, thus the memory efficiency of DCSS provides competitive merit.

\subsection{Image Super Resolution on DIV2K}
\textbf{Training and searching settings}.
We test DCSS on super resolution task with EDSR \cite{lim2017enhanced} being the base model.
The operating network is the single-scale $\times{}2$ model which is comprised of 16 residual blocks with 64 channels.
Differing from the previous tasks, the mean absolute error (MAE) is the loss function for this task.
The network is trained for $120$ epochs with hyperparameters the same as \cite{lim2017enhanced}.

\begin{table}[htbp]
	\centering
	\begin{tabular}{c c c c }
		\toprule[1pt]
		Method & Pr.\;(\%) & PSNR\;(dB) & SSIM \\
		\toprule
		Baseline & $-$ & $34.51$ & $0.9427$ \\
		\midrule
		Uniform & $48.3$ & $34.45$ & $0.9421$ \\
		DCSS    & $49.8$ & $34.45$ & $0.9422$ \\
		\midrule
		Uniform & $73.4$ & $34.25$ & $0.9406$ \\
		DCSS    & $74.3$ & $34.32$ & $0.9414$ \\		
		\bottomrule[1pt]
	\end{tabular}
	\caption{Comparison between DCSS and uniform pruning for EDSR. "Uniform" denotes uniform pruning.}
	\label{tab:8}
\end{table}

\textbf{Performance study and comparison}.
Super resolution is a memory-consuming regression problem which is very different from classification.
We display the results in Table \ref{tab:8}.
We still compare DCSS with uniform pruning with prune ratio around $50\%$ and $75\%$.
It is worth noting that the sparsity combination given by DCSS is rather close to uniform pruning.
However, the difference may look small, it still matters.
When sparsity is around $50\%$, DCSS yields slightly better result than uniform pruning.
As the prune ratio rising to around $75\%$, DCSS gains larger advantage toward uniform pruning, judging by PSNR and SSIM. 

The searched combination of sparsities usually varies a lot layer by layer for classification.
Comparing to the case of super resolution, which yields a combination close to uniform with small alteration, the searched combinations are totally different.
It provides clear evidence that task-specific searching is the better choice than transferring models pruned under different tasks.
Once again, the results prove the necessity of task-specific searching and the effectiveness of DCSS.

\section{Conclusion}
We introduce a novel differentiable search algorithm, named as DCSS, for network pruning. 
Comparing with other methods in this category, our approach keeps more information in feature maps and is more memory efficient during searching, thanks to the weight sharing within filters technique. 
To fully examine DCSS, we conduct experiments on three tasks, including image classification, semantic segmentation and super resolution.
For image classification, DCSS achieves the state-of-the-art performance on both CIFAR-100 and ImageNet.
As shown in Fig.\;\ref{fig:resnet18}, 
DCSS offers high stability and outperforms other pruning approaches on ImageNet consistently.
Low memory consumption during searching enables direct execution of DCSS on segmentation and super resolution.
Results of the experiments validate the necessity of task-specific searching as DCSS yields completely different patterns of sparsity combination for different tasks and outperforms other options like model transferring and uniform pruning.
For future work, we would like to extend our approach to more complicated operators such as channel attention.

\section*{Acknowledgment}
The authors would like to acknowledge Mr. Jingwei Chen for providing valuable and instructive suggestions in the field of model compression when he worked at Hisilicon.

\bibliographystyle{IEEEtrans}
\bibliography{refs}

\end{document}